\documentclass[pdflatex,sn-nature]{sn-jnl}

\usepackage{graphicx}%
\usepackage{multirow}%
\usepackage{amsmath,amssymb,amsfonts}%
\usepackage{amsthm}%
\usepackage{mathrsfs}%
\usepackage[title]{appendix}%
\usepackage{xcolor}%
\usepackage{textcomp}%

\usepackage{manyfoot}%
\usepackage{booktabs}%
\usepackage{tabularx}%
\usepackage{array}%
\newcolumntype{Y}{>{\raggedright\arraybackslash}X}%
\makeatletter
\renewcommand{\tablebodyfont}{\reset@font\fontsize{9bp}{11bp}\selectfont}%
\renewcommand{\tablecolheadfont}{\reset@font\fontsize{9bp}{11bp}\selectfont\bfseries\boldmath}%
\renewcommand{\tablecaptionfont}{\reset@font\fontsize{9bp}{11bp}\selectfont}%
\makeatother
\usepackage{algorithm}%
\usepackage{algorithmicx}%
\usepackage{algpseudocode}%
\usepackage{listings}%

\theoremstyle{thmstyleone}%

\theoremstyle{thmstyletwo}%

\theoremstyle{thmstylethree}%

\begin{document}

\title{Grower-in-the-loop AI for Deployable Precision Fertilisation in Sustainable Agriculture}

\author[1]{\fnm{Xulong} \sur{Wang}}\email{xl.wang@sheffield.ac.uk}

\author[2]{\fnm{Yu} \sur{Zhang}}\email{yuzhang1@zjut.edu.cn}

\author[1]{\fnm{Gaoshan} \sur{Bi}}\email{gaoshan.bi@sheffield.ac.uk}

\author[1]{\fnm{Menghui} \sur{Zhou}}\email{menghui.zhou@sheffield.ac.uk}

\author[3]{\fnm{Simon} \sur{Jeffery}}\email{sjeffery@harper-adams.ac.uk}

\author[3]{\fnm{Max} \sur{Davis}}\email{mdavis@harper-adams.ac.uk}

\author[4]{\fnm{Sam} \sur{Maycock}}\email{smaycock@velcourt.co.uk}

\author[5]{\fnm{Paul} \sur{Harris}}\email{paul.harris@rothamsted.ac.uk}

\author*[1]{\fnm{Po} \sur{Yang}}\email{po.yang@sheffield.ac.uk}

\affil[1]{\orgdiv{School of Computer Science}, \orgname{University of Sheffield}, \orgaddress{\city{Sheffield}, \country{UK}}}

\affil[2]{\orgdiv{College of Computer Science and Technology}, \orgname{Zhejiang University of Technology}, \orgaddress{\city{Hangzhou}, \country{China}}}

\affil[3]{\orgdiv{Centre for Crop and Environmental Science}, \orgname{Harper Adams University}, \orgaddress{\city{Newport}, \country{UK}}}

\affil[4]{\orgdiv{Trials Department}, \orgname{Velcourt Limited}, \orgaddress{\city{Stamford}, \country{UK}}}

\affil[5]{\orgname{Rothamsted Research}, \orgaddress{\street{North Wyke}, \city{Okehampton}, \country{UK}}}

\abstract{
In agriculture, typically over half of the applied nitrogen fertiliser is lost to water and to air.
Precision fertilisation aims to reduce these losses while maintaining yield and quality, yet existing tools struggle to capture farm-specific challenges in determining optimal nitrogen rates.
Static recommendations are common and overlook on-farm constraints, such as delays due to weather or machinery availability, or change in budgets. Each constraint governs when and how growers apply fertiliser in practice.
Here we present a grower-in-the-loop framework for winter wheat in England, combining baseline fertilisation planning from a machine learning-derived nitrogen-response curve with in-season adaptation via a large language model.
We evaluated the framework using data from long-term experiments (LTEs), replicated field trials, and farm deployments.
Retrospective model validation using LTEs confirmed low economic decision-loss across multiple seasons. Using field and farm trials, baseline schedules were shown to match standard-practice yields with less nitrogen, whereas in-season adjusted schedules additionally retained grain quality standards.
Given this enhanced performance, in-season adaptation absorbed weather, machinery and budget constraints to keep recommendations executable.
The framework therefore shows promise for translating static recommendations into adaptable, grower-informed decisions for sustainable fertiliser management.
}

\maketitle

Raising cereal yields through fertiliser use has long been a shared goal of growers, policymakers and researchers \cite{vanGrinsven2022nitrogen}.
However, excess application, especially of nitrogen (N), can cause nitrate enrichment of water bodies, nitrous oxide emissions to air and biodiversity loss through atmospheric deposition \cite{Gong2024reactive,vanGrinsven2022nitrogen}.
Further, more than half of applied N is typically lost, while N fertiliser synthesis and use together generate around 2\% of global greenhouse gas (GHG) emissions \cite{Galloway2008transformation,Menegat2022greenhouse}.
In response, the European Union's `Farm to Fork' strategy targets at least 50\% lower nutrient losses by 2030, which would reduce fertiliser use by at least 20\% \cite{EU2020FarmFork}.
In England, a statutory target requires agricultural loads of N, phosphorus and sediment entering water to fall at least 40\% below the 2018 baseline by 2038 \cite{UK2023EnvTargetsWater}.
Precision fertilisation is promoted as a primary means of meeting such targets \cite{vanGrinsven2022nitrogen,Xiao2024spatiotemporal}, optimising fertiliser use to balance crop yield, grain quality and GHG emissions at the field scale.
Intelligent solutions for sustainable fertiliser management (SFM) are therefore increasingly needed to help growers achieve this balance while improving the resilience of their farms to variable weather.

Wheat offers considerable scope for such solutions in England, covering 1,525 of 2,484 thousand ha of cereals in 2025 \cite{defra2026cerealareas}, and its optimum N rate varies substantially between seasons.
Yet, the UK Nutrient Management Guide (RB209) recommends N rates from fixed soil and rainfall classes and does not adapt to the season in progress \cite{AHDB2021RB209,Mi2025benchmark}.
For example, a study of six commercial winter wheat fields in England (2010--2012) found that RB209 rates varied little between fields, whereas economically optimal N rates varied widely \cite{Kindred2016autoN}.
Specifically, RB209 recommended a narrow range of 190 to 210 kg N/ha across these six fields, whereas the measured optima spanned more widely from 10 to 322 kg N/ha, exposing growers to substantial risks of both over- and under-fertilisation, and consequences thereafter.
Despite policy pressure to reduce N use, wheat of milling quality typically requires 13\% grain protein, which rises with N supply, so reducing N may cause revenue loss if grain misses this specification \cite{Hawkesford2014,AHDB2026enduse}.
Adaptive SFM tools are therefore urgently needed to guide site-specific N recommendations that dynamically respond to the season's weather, helping growers to balance yield and grain protein with N losses.

Many existing SFM tools, from empirical guides such as RB209 to process-based and machine learning (ML) models, primarily estimate agronomically optimal N rates and remain fixed once issued \cite{AHDB2021RB209,Keating2003APSIM,Jones2003DSSAT,Xiao2024spatiotemporal,Ruan2024datadriven}.
However, they rarely consider how recommendations are implemented and adjusted under changing on-farm constraints \cite{Rose2016decision,GAO2024precision}.
In practice, weather delays, Nitrate Vulnerable Zone closed periods \cite{UK2015NVZ}, machinery limits and budget revisions frequently prevent recommended schedules from being executed as advised.
In February 2026, for instance, England received 170\% of its long-term average rainfall, saturating soils \cite{EA2026WaterFeb} and restricting field access across Yorkshire and the North East through to March of that year \cite{AHDB2026cropMarch}.
Managing such disruptions relies on growers' practical knowledge of field access, machinery capacity and labour windows, which existing guides or models cannot readily capture \cite{Kloppenburg1991social,Nonaka1994dynamic}.
Consequently, growers routinely revise application timings and rates, yet such experience-based revisions remain unverified and are unlikely to preserve precision fertilisation optimality.
Capturing a grower's inherent knowhow demands bringing them into the SFM as active participants to better balance farm profitability with national targets for reducing N losses.

\begin{figure}[htbp]
\centering
\includegraphics[width=1\textwidth]{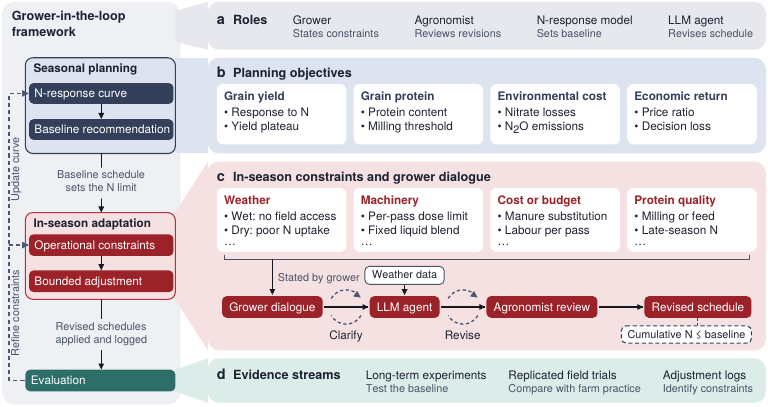}
\caption{\textbf{Overview of the grower-in-the-loop framework.} In planning, a fixed seasonal N recommendation is set from the N-response model against four agronomic objectives. In-season adaptation, operational constraints stated by the grower are turned into bounded adjustments by an LLM, and each adjustment is approved by an agronomist. Field evaluation draws on long-term experiments (LTEs) together with field-trial plots (at research and commercial farms) and commercial farms for wheat production. Dashed arrows mark in-season revision.}\label{fig:framework}
\end{figure}

Here, a grower-in-the-loop decision-support framework is presented that adapts fertiliser recommendations to real-world farming constraints in winter wheat (Fig.~\ref{fig:framework}).
Implemented within the Parallel Farm platform\footnote{\url{https://www.mutus.co.uk/fertiliser/}}, the framework integrates seasonal planning with in-season adaptation. 
For seasonal planning at sowing, it establishes a baseline N recommendation from an ML-derived N-response curve, calibrated to minimise economic decision loss. 
For in-season adaptation, a large language model (LLM)-based agent translates grower-reported operational constraints and current weather conditions into revised application schedules \cite{Tzachor2023LLM,Yi2025AgriLLM}. 
To mitigate hallucinations and numerical errors associated with LLMs \cite{Ji2023hallucination,Huang2025hallucination,Frieder2023numerical}, proposed adjustments are constrained so that cumulative N application cannot exceed the baseline seasonal recommendation and are reviewed by an agronomist before delivery.
The framework is evaluated using three complementary evidence streams, each with sites growing wheat. 
First, the baseline recommendation was validated retrospectively against data from two long-term experiments (LTEs), spanning 53 and 43 seasons, respectively, to assess its economic decision loss  (Fig.~\ref{fig:pre-validation} and Methods).
Second, replicated field trials across research and commercial sites, over three growing seasons, were used to compare an earlier recommendation model (2023--2024), static baseline recommendations (2024--2025) and LLM-adjusted recommendations (2025--2026) against standard farm practice, focusing on N use, grain yield and protein content (Figs.~\ref{fig:post-field-trials} and~\ref{fig:post-adjusted-trials}).
Third, in-season adjustments and their underlying causes were logged across two large-scale commercial farms and two managed field-trial scenarios in 2025--2026, to identify the operational constraints that can prompt real-world modifications (Table~\ref{tab:in-triggers}).
Together, these evidence streams demonstrate the feasibility of translating static fertiliser recommendations into dynamic, grower-informed decisions, with the potential to reduce fertiliser inputs without incurring significant penalties in yield.

\section{Results}

\subsection{Pre-season: Formulating Recommendations at Sowing Using LTEs}\label{sec:pre-season}

\begin{figure}[!htbp]
\centering
\includegraphics[width=\textwidth]{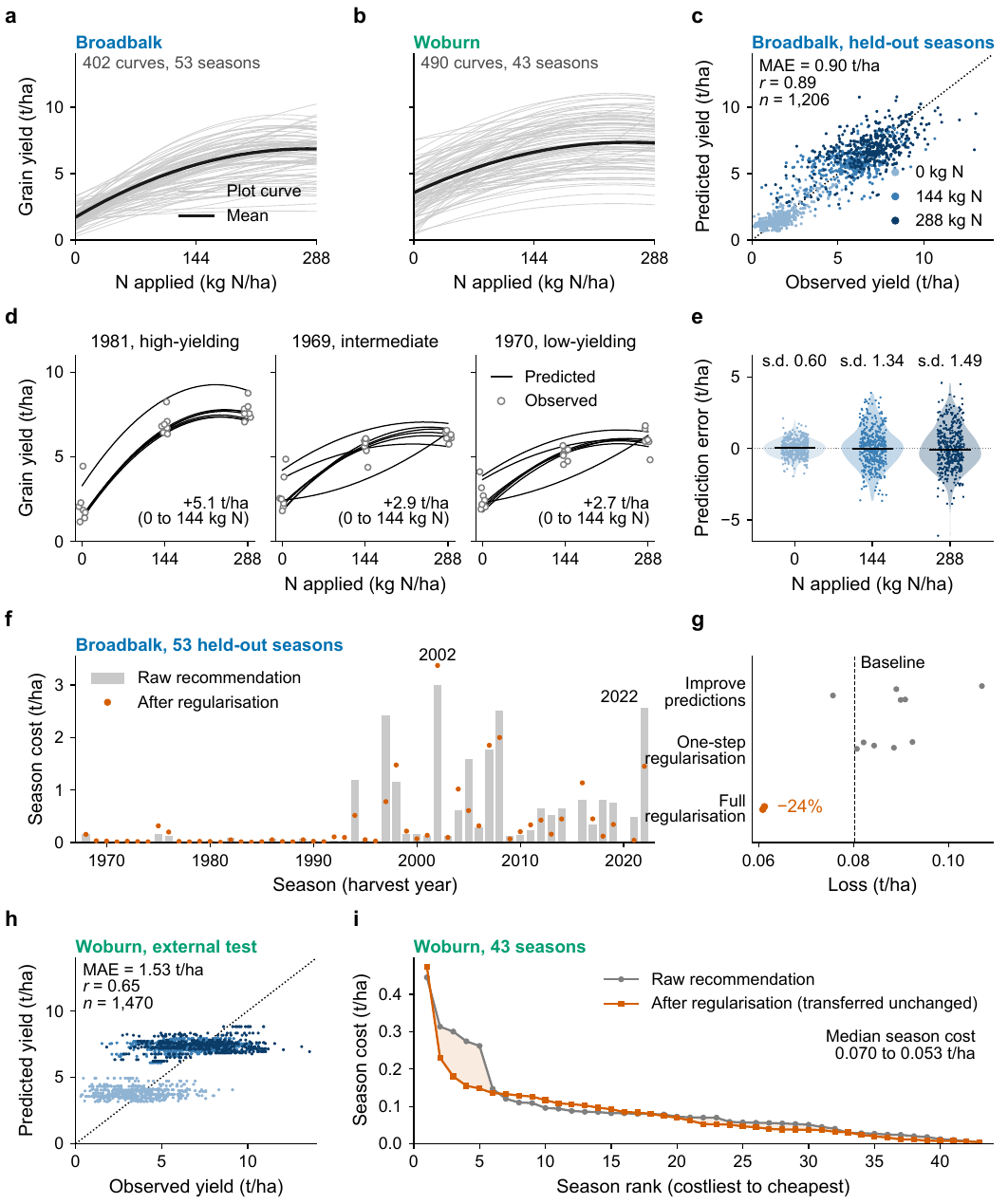}
\caption{\textbf{Validation of the N-response model and its rate regularisation on two long-term experiments (LTEs).}
\textbf{a,b}, Observed N-response curves, one per section per season, at the `Broadbalk' LTE (\textbf{a}, 402 curves, 53 seasons) and the `Woburn Ley-Arable' LTE (\textbf{b}, 490 curves, 43 seasons). Bold lines show the mean curve.
\textbf{c,h}, Predicted against observed anchor yields under leave-one-season-out validation at Broadbalk (\textbf{c}, $n = 1{,}206$) and under external test at Woburn (\textbf{h}, $n = 1{,}470$). Dotted lines are 1:1 lines. Point colour indicates N rate.
\textbf{d}, Predicted curves (lines) and held-out observations (points) in a high-yielding (1981), an intermediate (1969) and a low-yielding (1970) season.
\textbf{e}, Prediction error at 0, 144 and 288 kg N/ha. Violins show the error distribution and points show individual curves.
\textbf{f,i}, Season cost, that is, grain lost against the best possible N choice, before and after regularisation. Broadbalk seasons are shown in order (\textbf{f}, raw as bars, regularised as points) and Woburn seasons are ranked from costliest to cheapest (\textbf{i}).
\textbf{g}, Loss of the thirteen variants across the 53 held-out Broadbalk seasons, grouped by type of change (Extended Data Table~\ref{tab:pre-variants}). The dashed line marks the baseline. Season cost and loss are defined in Methods.}\label{fig:pre-validation}
\end{figure}

At sowing, the seasonal recommendation is derived from an N-response model built on the tabular foundation model TabPFN (Fig.~\ref{fig:framework} and Methods).
The model was validated on 892 N-response curves from two wheat-crop LTEs hosted at Rothamsted Research, `Broadbalk' and `Woburn Ley-Arable' (Fig.~\ref{fig:pre-validation}a,b and Methods).
At both sites, grain yield rose steeply over the first 144 kg N/ha and then levelled off (Fig.~\ref{fig:pre-validation}a,b).

Anchor yields, i.e. the yields at 0, 144 and 288 kg N/ha, were predicted under leave-one-season-out validation across the 53 Broadbalk seasons (Methods).
The mean absolute error (MAE) was 0.90 t/ha ($r = 0.89$, $n = 1{,}206$, Fig.~\ref{fig:pre-validation}c).
The predicted curves also followed the observed curves in seasons of contrasting yield (Fig.~\ref{fig:pre-validation}d).

However, the spread of the prediction error increased towards the high-N end of the curve (Fig.~\ref{fig:pre-validation}e).
The standard deviation of the error rose approximately 2.5-fold, from 0.60 t/ha at 0 kg N/ha to 1.49 t/ha at 288 kg N/ha.
Across the 53 held-out seasons, the median season cost of the raw recommendation was 0.13 t/ha (Fig.~\ref{fig:pre-validation}f and Methods).
In 2002 and 2022, the season cost rose to 2 to 3 t/ha, more than an order of magnitude above the median.

Thirteen variants of the recommendation were compared under one evaluation protocol (Fig.~\ref{fig:pre-validation}g, Extended Data Table~\ref{tab:pre-variants} and Methods).
Variants that changed the predictions, through new features, larger ensembles or conservative quantiles, reduced the loss by at most 6\% relative to the baseline.
Neither a shift alone nor a shrinkage alone reduced the loss below the baseline.
The full rate regularisation (Methods) reduced the median loss by 24\%, from 0.080 to 0.061 t/ha (Extended Data Table~\ref{tab:pre-variants}).
However, this reduction was not significant (season-paired Wilcoxon $P = 0.65$).
The reduction in season cost was concentrated in the costliest seasons (Fig.~\ref{fig:pre-validation}f).

The model and its rate regularisation were then applied, without recalibration, to the Woburn Ley-Arable LTE, which was used neither in training nor in calibration.
Anchor yields were predicted with a mean absolute error of 1.53 t/ha ($r = 0.65$, $n = 1{,}470$, Fig.~\ref{fig:pre-validation}h).
The regularisation lowered the median season cost by 24\%, from 0.070 to 0.053 t/ha (Fig.~\ref{fig:pre-validation}i).
Across all 43 seasons, the season cost fell by 22\%, and 24 seasons were improved (Wilcoxon $P = 0.79$).
However, the regularisation raised the season cost in the 1976 drought, from 0.26 to 0.47 t/ha.
At both sites, the most extreme seasons remained costly after regularisation (2002 and 2022 at Broadbalk, 1976 at Woburn).

\subsection{In-season: Grower-in-the-loop Adjustment at Commercial Sites}\label{sec:in-season}

In the 2025 to 2026 pilot season, the framework was deployed in four commercial scenarios (Fig.~\ref{fig:framework} and Methods).
These were a milling-wheat farm, a feed-wheat farm and two commercially run field trials, termed `C1' and `C2'.
Eighteen adjustment triggers were processed by the LLM layer (Table~\ref{tab:in-triggers}).
Seventeen were operational constraints in four categories, and none of the seventeen concerned the predicted N response.
The remaining trigger was an LLM proposal error, in which feed-wheat logic was applied to the milling scenario.
This proposal was intercepted at agronomist review and reverted within the same day.

\begin{table}[htbp]
\caption{Adjustment triggers and grower-in-the-loop responses in the pilot deployment season (2025 to 2026). Seventeen triggers denote operational constraints that the ML recommendation could not represent, and the single trigger in the category "other" is an LLM proposal error intercepted at review. Counts are aggregated across all four scenarios. Every response kept the seasonal N total at or below the amount defined by that recommendation.}\label{tab:in-triggers}
\setlength{\tabcolsep}{5pt}
\begin{tabularx}{\textwidth}{@{}l c c Y Y@{}}
\toprule
Constraint category & Triggers & Scenarios & Representative grower constraint & LLM response \\
\midrule
Weather delay    & 6 & 4 & Saturated soils prevented field access in February & Doses merged or deferred, seasonal total retained \\[4pt]
Machinery        & 4 & 4 & A single dose above 100 kg N/ha was unworkable, and sulphur was fixed within the liquid blend & Split sizes capped, and treated liquid urea with inhibitor adopted \\[4pt]
Cost or budget   & 4 & 2 & Manure and slurry replaced bagged nutrients, and a May dose was skipped on a sister field & Phosphorus, potassium and sulphur removed, ceiling lowered, dose skipped \\[4pt]
Protein quality  & 3 & 3 & The milling target required late-season nitrogen for grain protein & Flag-leaf nitrogen restored or retained \\[4pt]
Other            & 1 & 1 & Feed-wheat logic was misapplied to a milling scenario & Proposal reverted within the same day \\
\midrule
Total            & 18 & 4 & & \\
\bottomrule
\end{tabularx}
\end{table}

Cost or budget constraints were recorded only on the two commercial farms, and none were recorded at C1 or C2 (Table~\ref{tab:in-distribution}).

\begin{table}[htbp]
\centering
\caption{Distribution of the eighteen adjustment triggers by scenario and constraint category. The two large-scale commercial farm scenarios span all five categories. The two commercially run field-trial scenarios span only three categories and record no cost or budget trigger. The single "other" trigger is the intercepted LLM proposal error.}\label{tab:in-distribution}
\footnotesize
\begin{tabular}{lrrrrrr}
\toprule
Scenario & Weather & Machinery & Cost & Protein & Other & Total \\
\midrule
Commercial, milling & 2 & 1 & 1 & 1 & 1 & 6 \\
Commercial, feed    & 1 & 1 & 3 & 0 & 0 & 5 \\
Commercial, field trial, C1 & 1 & 1 & 0 & 1 & 0 & 3 \\
Commercial, field trial, C2 & 2 & 1 & 0 & 1 & 0 & 4 \\
\midrule
Total               & 6 & 4 & 4 & 3 & 1 & 18 \\
\bottomrule
\end{tabular}
\end{table}

In every scenario, N was redistributed within the seasonal total set by the recommendation (Methods).
The seasonal total was unchanged at the milling farm.
At the feed farm, it was reduced by 18.2\% after a soil-specific RB209 recalculation requested by the grower.
At C1 and C2, single large doses were split.

\subsection{Post-season: Recommendation against Standard Practice for All Field Trials}\label{sec:post-season}

In the 2024 to 2025 season, the recommendation issued at sowing was compared, unadjusted, with standard practice at five field trials (Methods).
These were the commercially run C1 and C2 trials, together with three trials at Harper Adams University, termed `H1', `H2' and `H3'.

\begin{figure}[!htbp]
\centering
\includegraphics[width=\textwidth]{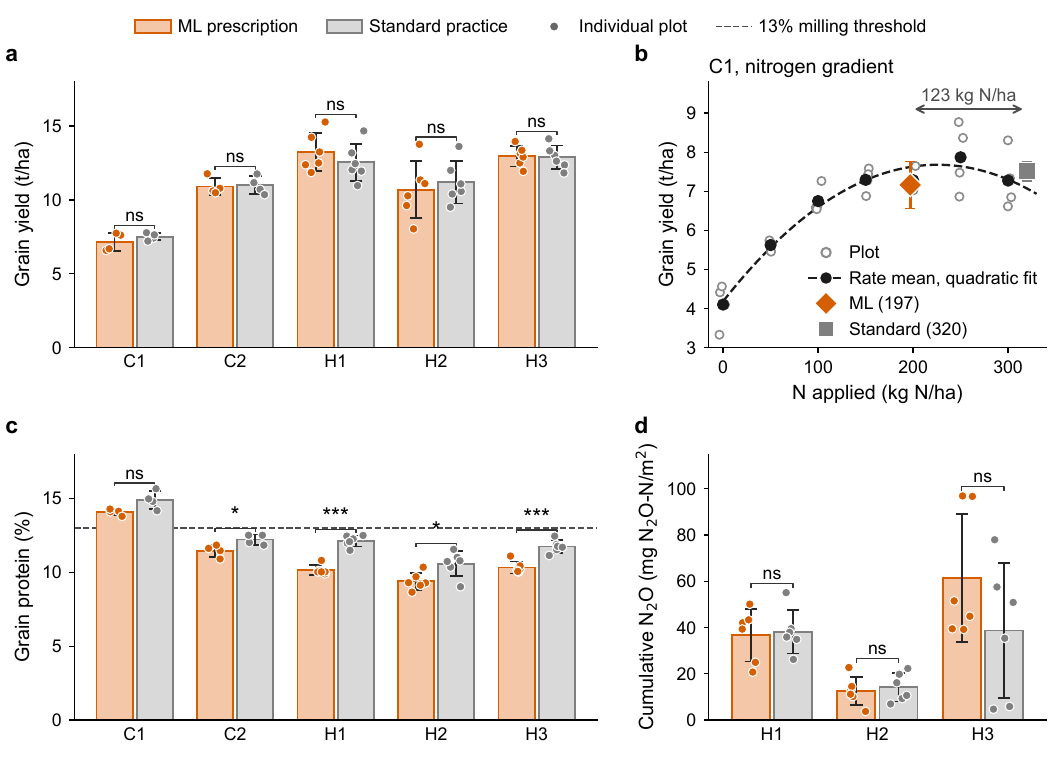}
\caption{\textbf{Field outcomes of the ML recommendation against agronomist standard practice at five sites.}
\textbf{a}, Grain yield at field trials C1, C2, H1, H2 and H3.
\textbf{b}, Nitrogen dose response at C1 across seven ammonium nitrate rates (0 to 300 kg N/ha). Open circles show individual plots, filled circles show rate means and the dashed line is a quadratic fit to the rate means. Operating points mark the recommendation (diamond, 197 kg N/ha) and standard practice (square, 320 kg N/ha).
\textbf{c}, Grain protein content at C1, C2, H1, H2 and H3. The dashed line marks the 13\% milling-wheat protein threshold \cite{Irisarri2026grassland}.
\textbf{d}, Cumulative N$_2$O emissions at H1, H2 and H3.
In \textbf{a}, \textbf{c} and \textbf{d}, bars show the treatment mean, error bars denote one standard deviation and points show individual plots. Sample sizes were $n = 4$ plots per treatment at C1 and C2 and $n = 6$ at H1, H2 and H3. Symbols denote treatment comparisons (ns not significant, $\ast$ $P < 0.05$, $\ast\ast\ast$ $P < 0.001$), tested with a Welch t-test at C1 and C2 and a one-way ANOVA at H1, H2 and H3. The recommendation is the seasonal nitrogen amount and timing set by the ML model at sowing, and standard practice is the seasonal nitrogen amount and timing set by the agronomist. No in-season adjustment layer was active.}\label{fig:post-field-trials}
\end{figure}

There was no significant difference in grain yield between the recommendation and standard practice at any of the five trial sites (Fig.~\ref{fig:post-field-trials}a).
Yields were 7.16 and 7.52 t/ha at C1 (Welch $P = 0.34$) and 10.91 and 11.00 t/ha at C2 (Welch $P = 0.84$).
At H1, H2 and H3, one-way ANOVA gave $P = 0.363$, $P = 0.618$ and $P = 0.852$, respectively.

At C1, the N-rate ladder showed a yield plateau above approximately 150 kg N/ha (Fig.~\ref{fig:post-field-trials}b).
The recommendation was applied at 197 kg N/ha, within the plateau, and standard practice at 320 kg N/ha.
The additional 123 kg N/ha under standard practice gave no significant yield gain (0.36 t/ha, Welch $P = 0.34$).

Grain protein content was lower under the recommendation at all five sites (Fig.~\ref{fig:post-field-trials}c).
At H1, H2 and H3, protein fell from 12.12 to 10.15\%, from 10.58 to 9.40\% and from 11.75 to 10.33\%, respectively.
Each of these reductions was significant (one-way ANOVA, $P < 0.0001$, $P = 0.019$ and $P = 0.0002$).
At C2, protein fell from 12.22 to 11.46\% (Welch $P = 0.03$).
At C1, protein fell from 14.90 to 14.08\%, but the reduction was not significant (Welch $P = 0.07$).
Trial C1 was the only site above the 13\% milling-wheat protein threshold, under both treatments.

Cumulative N$_2$O emissions were measured at H1, H2 and H3 only (Fig.~\ref{fig:post-field-trials}d).
There was no significant difference between treatments at any of the three sites (one-way ANOVA, $P = 0.81$, $P = 0.66$ and $P = 0.20$).
At H1 and H2, emissions were slightly lower under the recommendation, at 36.8 against 38.3 and 12.6 against 14.2 mg N$_2$O-N/m$^2$, respectively.
However, at H3, emissions were higher under the recommendation, at 61.5 against 38.7 mg N$_2$O-N/m$^2$.
Standard deviations at H3 were approximately 28 mg N$_2$O-N/m$^2$ in both groups.

\subsection{Post-season: Adjusted Recommendation against Standard Practice for Commercial Field Trials}\label{sec:post-adjusted}

In the 2025 to 2026 season, the adjusted recommendation, i.e. the recommendation after grower-in-the-loop adjustment, was compared with standard practice at trial sites C1 and C2 (Methods).

\begin{figure}[!htbp]
\centering
\includegraphics[width=\textwidth]{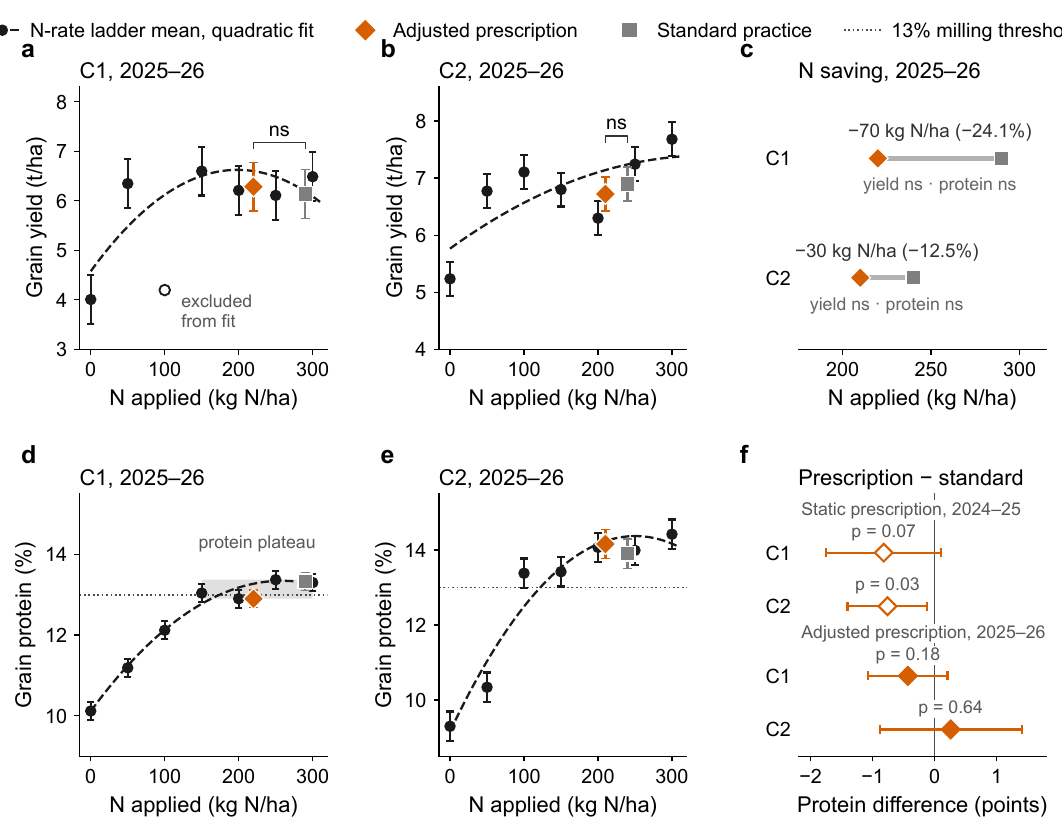}
\caption{\textbf{Field outcomes of the adjusted recommendation against standard practice in the 2025 to 2026 pilot season.}
\textbf{a},\textbf{b}, Grain yield against nitrogen applied at field trials C1 (\textbf{a}) and C2 (\textbf{b}).
\textbf{c}, Seasonal nitrogen applied under the adjusted recommendation and standard practice at C1 and C2.
\textbf{d},\textbf{e}, Grain protein content against nitrogen applied at C1 (\textbf{d}) and C2 (\textbf{e}). The dotted line marks the 13\% milling-wheat protein threshold, and the shaded band in \textbf{d} spans the ladder means from 150 to 300 kg N/ha.
\textbf{f}, Difference in grain protein content between the recommendation and standard practice, for the static recommendation in 2024 to 2025 (open diamonds) and the adjusted recommendation in 2025 to 2026 (filled diamonds).
In \textbf{a},\textbf{b},\textbf{d},\textbf{e}, filled circles show the means of seven ammonium nitrate rates and the dashed line is a quadratic fit to these means. The open circle in \textbf{a} is the 100 kg N/ha mean, excluded from the fit. Error bars denote one standard error of the mean, derived from the least significant difference of each trial with $n = 4$ plots per treatment. In \textbf{f}, error bars denote the 95\% confidence interval. The interval was derived from a Welch t-test on plot data for 2024 to 2025 and from the pooled error of each trial for 2025 to 2026. Symbols in \textbf{a},\textbf{b} denote the treatment comparison (ns, not significant), tested with a t-test on the pooled error of each trial. The adjusted recommendation is the recommendation after grower-in-the-loop adjustment (Table~\ref{tab:in-triggers}).}\label{fig:post-adjusted-trials}
\end{figure}

Without N, grain yield was 4.01 t/ha at C1 and 5.24 t/ha at C2, i.e. 61\% and 68\% of the highest ladder yield (Fig.~\ref{fig:post-adjusted-trials}a,b).
The first 50 kg N/ha increased yield by 2.34 t/ha at C1 and by 1.54 t/ha at C2.
At C1, every rate from 50 kg N/ha upward lay within one least significant difference of the highest ladder yield.
At C2, the response was irregular, and the highest yield of 7.68 t/ha was reached at 300 kg N/ha.

Grain protein increased with N, by 1.05 percentage points per 100 kg N at C1 ($P = 0.003$) and by 1.67 points at C2 ($P = 0.007$).
Protein reached the 13\% milling threshold at approximately 148 kg N/ha at C1 and 94 kg N/ha at C2 (Fig.~\ref{fig:post-adjusted-trials}d,e).
At C1, protein remained between 12.9 and 13.4\% across ladder rates from 150 to 300 kg N/ha.

The adjusted recommendation applied less N than standard practice at both trials (Fig.~\ref{fig:post-adjusted-trials}c).
At C1, it applied 220 against 290 kg N/ha, i.e. 70 kg N/ha or 24.1\% less.
At C2, it applied 210 against 240 kg N/ha, i.e. 30 kg N/ha or 12.5\% less.
There was no significant difference in grain yield, at 6.28 against 6.14 t/ha at C1 ($P = 0.83$) and 6.72 against 6.90 t/ha at C2 ($P = 0.68$).
Both operating points lay on the yield plateau of their field (Fig.~\ref{fig:post-adjusted-trials}a,b).
Confidence intervals were wide, and formal equivalence was not tested.

There was no significant difference in grain protein between the adjusted recommendation and standard practice at either trial (Fig.~\ref{fig:post-adjusted-trials}d,e).
At C2, protein was 14.16\% under the adjusted recommendation and 13.90\% under standard practice ($P = 0.64$), and both exceeded the milling threshold.
At C1, protein was 12.90 and 13.33\%, respectively ($P = 0.18$), although standard practice applied 70 kg N/ha more.

The protein difference from standard practice was smaller than under the unadjusted recommendation in 2024 to 2025 (Fig.~\ref{fig:post-adjusted-trials}f).
In 2025 to 2026, the differences were $-0.43$ points at C1 and $+0.26$ points at C2, and neither was significant.
The N reduction was also smaller than in 2024 to 2025, at 24.1 against 38.4\% at C1 and 12.5 against 17.6\% at C2.
Under the adjusted recommendation, a late dose of 44.6 and 40.0 kg N/ha was retained at C1 and C2, respectively.
One protein-quality trigger was logged at each trial (Table~\ref{tab:in-distribution}).
At C1, the protein change per 100 kg N saved was similar in both seasons, at 0.67 and 0.61 percentage points.
The trials were on different fields in the two seasons, with a different standard practice, so this comparison is observational.

\section{Discussion}\label{sec:discussion}

This study asked whether the efficacy of a deployed N recommendation should be set by prediction accuracy, or by a decision layer combined with in-season operability.
The three evidence streams presented support the second answer, although the field and deployment evidence remains at a pilot scale.

In retrospective validation on two LTEs, season cost was reduced by the decision rule rather than by sharper predictions.
Prediction error was largest at high N rates, where the profit optimum lies.
Near the optimum, a small error in the slope of the curve can shift the recommended rate by tens of kilograms.
Shrinking each curve towards a long-run section average likely regularised the decision against this error.
The same rate regularisation transferred from the Broadbalk LTE to the Woburn Ley-Arable LTE without recalibration, possibly because it encodes no site knowledge.
This agrees with de Lara et al.~\cite{deLara2023EONR}, who found that tools of similar accuracy can recommend different optimum rates.

At the four commercial sites (two farms and two field trials, C1 and C2) in the pilot season, every constraint-driven change concerned the execution of the recommendation, and none concerned the predicted N response.
Weather windows, machinery limits and budget revisions arise after sowing and are known first to the grower, so a recommendation fixed at sowing cannot represent them \cite{Kloppenburg1991social,Nonaka1994dynamic}.
Cost or budget constraints were recorded only on the two commercial farms, possibly because the input budget of a field trial is fixed before the season starts.
Earlier LLM crop-management systems were tested in simulators or on retrospective records \cite{Wu2024agronomists,Chen2025RLLLM,Balasubramaniam2026AgriSAGE,Chamara2026ChatGPT,Tzachor2023LLM,Yi2025AgriLLM}.
Here, growers stated constraints during a live season, and these were translated into bounded dose changes.
On review, one unsound LLM proposal was intercepted, so the review step remains necessary \cite{Ji2023hallucination,Frieder2023numerical}.
These event-level records agree with adoption research on operational barriers and on the preference of advisors for human oversight \cite{McCown2002changing,GAO2024precision,Adereti2023barriers,Gardezi2026advisors}.

Over three seasons of field trials, the recommendation changed from an N total to the static recommendation, and then the adjusted recommendation (Extended Data Figs.~\ref{fig:post-three-seasons-yield} and~\ref{fig:post-three-seasons-nue}).
In 2023 to 2024, only a seasonal N total was recommended at three commercially run field trials (C1, C2 and C3), and doses were applied on the standard-practice dates.
The recommendation applied 18\% more N without a yield gain, so yield per unit of N fell at all three commercial trials.
Yield at C1 was also significantly lower, for an unrecorded reason.
From 2024 to 2026, the recommendation applied 21\% less N than standard practice on average, with no significant yield loss at any of the seven site-seasons (two farms and five trials).
Under the static (unadjusted) recommendation, grain protein was lower at all five trials (C1, C2, H1, H2, H3), likely because a lower N total reduces the N available during grain fill.
Under the adjusted recommendation (for C1 and C2 only), protein did not differ significantly at C1 or C2, possibly owing to the retained late dose or the smaller N reduction.
Pooled over 2024 to 2026, grain protein content remained 0.9 percentage points lower ($P = 0.016$), so the recovery holds for the adjusted recommendation only.

Some limitations remain.
First, both LTEs lie in SE England and share one regional climate.
At neither site was the reduction in season cost statistically significant, and the most extreme seasons remained costly after regularisation.
The transfer of such outcomes to other locations with different climates remains to be tested, say with locally relevant LTEs or (better still) local on-farm rate trials.
Second, H1--H3 were single small-plot trials conducted on a different field each year, and deployment was limited to a single season.
A multi-season, statistically designed trial on commercial fields is therefore required before trial-scale N reductions can be expected in practice \cite{EU2020FarmFork}.
Third, N$_2$O emissions were measured at three study trials in one season only, where emissions did not differ significantly between treatments and varied widely within sites.
Given such limited information, it is not possible to conclude that the lower N inputs reduced emissions.

Overall, this study's evidence suggests that precision fertilisation is limited by operability as much as by prediction accuracy.
In a pilot season, a bounded LLM layer had the grower as the authoritative decision maker and the agronomist as a safety layer.
It absorbed the constraints that arose after sowing, while the seasonal N total held.
Grower-in-the-loop deployment, long argued for in adoption research \cite{McCown2002changing}, therefore appears to be a first-order design requirement for precision fertilisation within an AI framework.
The framework is intended to complement RB209, the national N use guide for the UK, since a well-applied RB209 recommendation remains competitive at typical price ratios \cite{Wang2026profit,AHDB2021RB209}.
We also suggest that for robust evaluations of N guidance tools in the long-term, on-farm records of planned against executed doses should be captured to build a valuable data resource over time.

\section{Methods}\label{sec:method}

A grower-in-the-loop framework is developed for winter wheat (Fig.~\ref{fig:framework}).
At sowing, a seasonal nitrogen recommendation is set by an N-response model, and rate regularisation is applied when its predicted curve is converted into an N rate.
In season, this recommendation is adjusted by an LLM layer, and each adjustment is reviewed by an agronomist.
The model and its rate regularisation are verified on two LTEs, and the regularised recommendation is tested against expert practice in field trials.
The coupled framework is then deployed in a pilot season.
Each component is described below in the order in which it enters the framework.
The framework is demonstrated for winter wheat in England, which ranks among the most nitrogen-intensive cereals grown at commercial scale in temperate systems \cite{FAO2023FAOSTAT}.
Nitrogen applications in England are constrained by regulatory windows, that is, the closed periods of Nitrate Vulnerable Zones, in addition to delays due to weather, delays due to machinery availability and changes in available budgets \cite{UK2015NVZ}.

\subsection{Long-term experiments}\label{sec:method-dataset}

Two LTEs of Rothamsted Research supply the N-response curves on which the recommendation model is built and tested.
Both are published by the Electronic Rothamsted Archive under a CC-BY 4.0 licence.
The Broadbalk Wheat LTE at Rothamsted (Hertfordshire, silty clay loam soils) provides annual grain yields for harvest years 1968 to 2022 \cite{Glendining2023Broadbalk}.
The seasons 2015 and 2020 are absent from the source data, leaving 53 seasons.
One N-response curve is fitted per section per season across the ten sections, that is, 402 curves in total, between six and eight per season.
For 280 curves, seven N rates are available (0, 48, 96, 144, 192, 240 and 288 kg N/ha).
For the 122 curves from seasons before the 240 and 288 kg N/ha rates were introduced, five rates from 0 to 192 kg N/ha are available.
For these curves, the yield at 288 kg N/ha is an extrapolation of the fitted curve and is flagged as such.
The Woburn Ley-Arable LTE (Bedfordshire, sandy loam soils) provides yields of winter wheat as the first test crop for harvest years 1976 to 2018 \cite{Glendining2022WoburnWheat}.
Of 2,240 raw records, 2,228 with a numeric grain yield were retained, and previous crops were decoded from the published cropping sequence \cite{Glendining2022WoburnCropping}.
The curve unit is the combination of season, block, rotation and previous farmyard manure group with at least four distinct N rates.
Curves were retained where the fit reached a coefficient of determination of at least 0.3 and the yield at 288 kg N/ha was not below that at 144 kg N/ha.
Curves whose previous crop could not be fully decoded were excluded, which resulted in 490 curves over all 43 seasons.
Together, the LTEs supply 892 curves from two sites that differ in soil texture and cropping history but share the same regional climate.
Weather covariates were taken from the HadUK-Grid 1 km dataset at the grid point of each LTE, that is, 51.82$^\circ$~N, 0.37$^\circ$~W for Broadbalk and 52.00$^\circ$~N, 0.62$^\circ$~W for Woburn Ley-Arable \cite{MetOffice2025HadUK}.
All weather windows use whole-month boundaries and are aligned to the harvest year, so, for example, the winter window for the harvest of 1996 runs from October 1995 to February 1996 (Extended Data Table~\ref{tab:pre-features}).

\subsection{N-response curves}\label{sec:method-curve}

Each plot-season N-response is described by a three-parameter curve combining exponential saturation with a linear term,
\begin{equation}
Y(N) = a\,r^{N} + b\,(1 - r^{N}) + c\,N,
\label{eq:curve}
\end{equation}
where $Y$ is grain yield (t/ha) and $N$ is the N rate (kg N/ha).
The parameter $a$ is the yield at zero N (t/ha) and $b$ is the plateau of the saturating term (t/ha).
The parameter $c$ is the slope of the linear term (t/ha per kg N/ha) and $r$ is a dimensionless shape parameter.
The shape parameter is held fixed at $r = 0.9945$ for every curve, so that three parameters remain to be determined per curve.
The Woburn Ley-Arable LTE curves were fitted by constrained least squares with the same form and the same $r$, under the constraint $\mathrm{d}Y/\mathrm{d}N \geq 0$ on 0 to 250 kg N/ha.
Given $r$, the yields at any three N rates determine the curve uniquely, because $(a, b, c)$ then solve a $3 \times 3$ linear system.
We take these three rates as 0, 144 and 288 kg N/ha, that is, the lowest, the central and the highest rate applied at the Broadbalk LTE.
This choice covers both ends and the centre of the tested range, at rates that were applied in the experiment rather than interpolated.
These three yields, $Y(0)$, $Y(144)$ and $Y(288)$, are termed the anchor yields below, because they pin the curve once $r$ is fixed.
Predicting a curve therefore reduces to predicting its three anchor yields, which are the prediction targets of the model and the quantities on which accuracy is evaluated (Extended Data Fig.~\ref{fig:method-nstar}a).
They number $402 \times 3 = 1{,}206$ at Broadbalk and $490 \times 3 = 1{,}470$ at Woburn.

\subsection{The recommendation model (TabPFN)}\label{sec:method-model}

The three anchor yields are predicted by TabPFN, a transformer-based tabular foundation model that performs in-context learning \cite{Hollmann2025TabPFN,Hollmann2023TabPFN}.
TabPFN is pre-trained on synthetic tabular tasks and predicts a new task by conditioning on its training rows at inference time.
It has been benchmarked against tuned gradient boosting across a wide range of small tabular datasets \cite{Hollmann2025TabPFN}.
As such, no gradient training is performed on the task data, and no hyperparameter search is run.
The packaged version 3 regressor weights distributed with the `tabpfn' Python package (version 8.1.0) were used.
Three independent regressions predict $Y(0)$, $Y(144)$ and $Y(288)$, and the curve is then solved through the anchors.
Ten input features are used, that is, three categorical descriptors of site and management, one section identifier and six weather variables (Extended Data Table~\ref{tab:pre-features}).
The N rate and the season are never used as features.
Site identity enters only through the section identifier, which takes values 0 to 9 at Broadbalk and the out-of-vocabulary code 999 for rows from any other site.
Categorical features are integer-encoded with a vocabulary built from the training fold only, and unseen categories are coded as $-1$.
The regressor was run with four estimators and a fixed random seed of 42, with all other settings at package defaults.
The predictive quantiles at 0.16 and 0.84 were requested for $Y(288)$ to drive the uncertainty guard described below.
Software was `tabpfn' 8.1.0 with PyTorch 2.5.1 on a single NVIDIA RTX 4060 Ti graphics processing unit with 16 GB of memory.
A full 53-fold evaluation of one model configuration takes about 200 s.

The recommended N rate is the profit maximiser of the predicted curve,
\begin{equation}
N^{*}(\rho) = \arg\max_{N \in [0, 600]} \left[ Y(N) - \frac{\rho}{1000}\,N \right],
\label{eq:argmax}
\end{equation}
where $N^{*}$ is the recommended rate (kg N/ha), $Y(N)$ is the predicted curve (t/ha) and $\rho$ is the fertiliser-to-grain price ratio (kg grain per kg N).
The maximisation is evaluated on a 1 kg grid.
We chose eight price scenarios, $\rho \in \{3, 4, 5, 6, 7, 8, 10, 12\}$, to span historical UK nitrogen-to-wheat price ratios and high-price stress cases.
The ratio $\rho$ is a deployment-time input.
Graphically, $N^{*}$ is the point at which a line of slope $\rho/1000$ touches the predicted curve, so a higher nitrogen price moves $N^{*}$ to a lower rate (Extended Data Fig.~\ref{fig:method-nstar}b).

\subsection{Validation protocol}\label{sec:method-validation}

Two benchmarks are used throughout, that is, leave-one-season-out validation at the Broadbalk LTE and a strict external test at the Woburn Ley-Arable LTE.
Accuracy at Broadbalk was assessed by leave-one-season-out validation over the 53 harvest seasons.
For each season, all curves of that season form the test set.
The curves of the remaining 52 seasons form the in-context training set.
Within each fold, three regressions are fitted, one for each anchor yield.
Every curve is therefore predicted exactly once by a model that never saw its season.
The procedure is deterministic, and repeated runs are bitwise identical under the fixed seed.
Woburn serves as a strict external test.
The model context is the full set of 402 Broadbalk curves, and every parameter of the regularisation is calibrated on Broadbalk only.
The value of this test rests on complete separation between the two sites, so that any transfer observed cannot be attributed to tuning.
No Woburn record enters training, calibration, threshold setting or model selection in any form.
At test time the section identifier is set to the out-of-vocabulary code for every Woburn row.
Mean absolute error (MAE) and the Pearson correlation coefficient ($r$) are computed between predicted and observed anchor yields, pooled over curves and the three anchor rates.
Observed anchor yields are the values of the fitted curve at 0, 144 and 288 kg N/ha, that is, the same labels used for modelling.
For the showcase seasons in Fig.~\ref{fig:pre-validation}d, seasons were ranked by mean observed yield and split into terciles.
Within each tercile, the season with the lowest season-level anchor error is shown, giving 1981, 1969 and 1970.
This is a display choice only, and Fig.~\ref{fig:pre-validation}c reports accuracy over all seasons.

\subsection{Rate regularisation}\label{sec:method-correction}

Let $\hat{\mathbf{y}}_i \in \mathbb{R}^{3}$ be the predicted anchor vector of curve $i$ and $s(i)$ its section.
The regularisation has three components applied in sequence.
First, the prediction is shrunk towards the long-run average of its section,
\begin{equation}
\tilde{\mathbf{y}}_i = \boldsymbol{\mu}_{s(i)} + \lambda_b \left( \hat{\mathbf{y}}_i - \boldsymbol{\mu}_{s(i)} \right),
\label{eq:shrink}
\end{equation}
where $\tilde{\mathbf{y}}_i$ is the regularised anchor vector (t/ha) and $\lambda_\rho$ is a dimensionless shrinkage factor chosen per price ratio $\rho$.
The vector $\boldsymbol{\mu}_{s(i)}$ is the mean predicted anchor vector of the same section in the calibration predictions (t/ha), taken from training seasons only.
The factor is searched over $\lambda_\rho \in \{0, 0.02, \ldots, 1\}$ with an increment of 0.02, and for an unknown section, as at the Woburn Ley-Arable LTE, the global mean replaces $\boldsymbol{\mu}_{s(i)}$.
Second, a price-specific shift $\delta_\rho \in \{-100, -98, \ldots, 100\}$ kg N/ha, with an increment of 2 kg N/ha, is added to the profit maximiser of the shrunk curve.
The result is clipped to 0 to 600 kg N/ha.
Third, an uncertainty guard halves $\lambda_\rho$ for curves whose predictive interval width for $Y(288)$ exceeds the 75th percentile of calibration widths.
Uncertain curves are thereby pulled twice as hard towards the average.
Calibration is nested.
Within the 52 training seasons of each outer fold, an inner leave-one-season-out produces out-of-fold calibration predictions.
The parameters $\lambda_\rho$ and $\delta_\rho$ are selected sequentially on those inner predictions ($\lambda_\rho$ first and then $\delta_\rho$ given $\lambda_\rho$) to minimise the median loss defined below.
No test-season information enters the calibration.

\subsection{Loss function and variant evaluation}\label{sec:method-loss}

The loss for curve $i$ at price ratio $\rho$ is the grain lost against the best possible N choice,
\begin{equation}
L_i(\rho) = \max_{N} \Pi_i(N) - \Pi_i\!\left(N_{\mathrm{rec},i}(\rho)\right), \qquad \Pi_i(N) = Y_i(N) - \frac{\rho}{1000}\,N,
\label{eq:regret}
\end{equation}
where $L_i(\rho)$ is the loss (t/ha) and $\Pi_i(N)$ is the profit of curve $i$ at rate $N$, expressed in grain (t/ha).
The curve $Y_i$ is the true fitted curve (t/ha) and $N_{\mathrm{rec},i}(\rho)$ is the recommended rate (kg N/ha).
The rate $N$ ranges over 0 to 600 kg N/ha on a 1 kg grid, and the loss is floored at zero.
The reported scalar is the mean over the eight price ratios of the median over curves, in t grain/ha per season.
The season cost shown in Fig.~\ref{fig:pre-validation}f is the within-season median of curve costs, each curve cost first averaged over the eight price ratios.
We scored thirteen variants under this identical evaluator and the same held-out seasons (Extended Data Table~\ref{tab:pre-variants}).
The variants fall into three groups, that is, five that improve the predictions themselves, five that adjust the recommendation in one step and two that apply the full regularisation.
The baseline feeds the predicted anchors directly to the profit maximiser.
This design separates a gain from sharper predictions from a gain from regularising the decision, under one evaluator and one set of seasons.

\subsection{Research and commercial field trials}\label{sec:method-trials}

Field trials were run in three seasons at winter wheat sites in England.
In 2023 to 2024, a seasonal N total of 212 kg N/ha, generated by an earlier version of the recommendation model, was recommended with no timing.
It was compared with standard practice (180 kg N/ha) at three commercially run sites, C1, C2 and C3.
Both treatments were applied in two equal doses on the standard-practice dates, 2 April and 20 May 2024.
In 2024 to 2025, the recommendation was compared with standard practice at five sites.
Two sites (C1 and C2) were managed by a commercial farm management company, and standard practice there was the company's expert N programme.
Three sites (H1, H2 and H3) were located at Harper Adams University, and standard practice there was an agronomist-led programme, with six replicate plots per treatment.
In this season the recommendation treatment received the seasonal N amount and split pattern generated at sowing by the TabPFN model with its rate regularisation, with no in-season adjustment.
In 2025 to 2026, the adjusted recommendation was compared with standard practice at C1 and C2 only.
The adjusted recommendation received the doses of the final adjusted recommendation (Section~\ref{sec:method-deployment}).
The trial at each of C1 and C2 moved to a different field of the same farm between seasons.

In 2023 to 2024, each trial comprised five treatments with four replicate plots: the recommendation, standard practice, an unfertilised control and two RB209 rates (110 and 220 kg N/ha).
In 2024 to 2025 and 2025 to 2026, each trial at C1 and C2 comprised nine treatments with four replicate plots each.
The nine treatments were the recommendation, standard practice and an N-rate ladder of ammonium nitrate at seven rates from 0 to 300 kg N/ha.
Ladder rates were split equally between the first two application timings, and the 0 kg N/ha rate served as the unfertilised control.
The ladder was used to locate the yield plateau and to describe the N response of each field (Figs.~\ref{fig:post-field-trials}b and~\ref{fig:post-adjusted-trials}).
Grain yield and grain protein content were measured at harvest, at all sites, across all years. Cumulative N$_2$O emissions were measured at H1, H2 and H3 through the 2024 to 2025 season, only.

Nitrogen use was summarised by two indices for each treatment.
Partial factor productivity (PFP) was grain yield divided by total N applied.
Agronomic efficiency (AE) was the yield gain over the unfertilised control divided by total N applied, and was calculated only where a 0 kg N/ha plot was included.
No site was replicated within a season, and the trials therefore bound what each recommendation delivers at the field scale.
Site codes follow an anonymisation key held by the authors.

\subsection{In-season deployment protocol}\label{sec:method-deployment}

The coupled framework was deployed in the 2025 to 2026 season across four scenarios.
These were two large-scale commercial farms (one growing milling wheat and one feed wheat) and the two commercial field-trial plots C1 and C2.
At sowing each scenario received a seasonal recommendation from the TabPFN model, specifying the total N amount and its split across application timings.
During the spring fertilisation window, a three-step grower-in-the-loop workflow was run for each scenario (Extended Data Fig.~\ref{fig:method-workflow}).
The workflow was placed within the five stages of the winter wheat season, from sowing in September to harvesting in August.
The recommendation was generated at sowing, whereas the workflow was confined to the fertilisation stage from February to May.
The workflow was triggered adaptively whenever the grower reported a change in field conditions or operational constraints.

First, the grower stated in natural language the current needs or the field operations planned for the coming weeks, and asked for advice on the recommendation.
Typical statements concerned feasible application dates, machinery limits, budget revisions and grain quality targets.
Second, the LLM layer, implemented with Gemini 3 Pro, received the current recommendation, the grower statement and a summary of the prevailing weather.
The weather summary was compiled for each cycle with the Deep Research function of Gemini.
The layer then proposed a revised split of doses.
Proposals were bounded to the seasonal total set by the recommendation, which could be retained or reduced but not exceeded, and to per-application dose limits stated by the grower.
The prompt templates used at this step are provided as Supplementary Information.
Third, an agronomist reviewed every proposal and confirmed, modified or rejected it before the adjusted recommendation was returned to the grower.
Each grower statement that led to a proposed change was logged as a trigger.
Triggers were assigned to one of five constraint categories, that is, weather delay, machinery, cost or budget, protein quality, or other.
Trigger counts, seasonal totals and review outcomes were compiled from these logs (Tables~\ref{tab:in-triggers} and~\ref{tab:in-distribution}).

\subsection{Statistics and reproducibility}\label{sec:method-statistics}

At the LTEs, the season was the experimental unit for paired tests.
Season costs with and without the rate regularisation were compared by a two-sided Wilcoxon signed-rank test, paired within season.
The test covered 53 seasons at the Broadbalk LTE and 43 seasons at the Woburn Ley-Arable LTE.
Uncertainty in the Broadbalk loss difference was estimated by a season-cluster bootstrap, with 10,000 resamples of seasons drawn with replacement.
The bootstrap gave a difference of 0.019 t/ha, with a 95\% interval that included zero.

For the 2023 to 2024 and 2025 to 2026 trials, treatment means and the least significant difference at $P = 0.05$ were taken from the analysis of variance of each trial.
In 2023 to 2024, a yield difference larger than the least significant difference was taken as significant, and no least significant difference was available for grain protein.
In 2025 to 2026, the recommendation was compared with standard practice by a t-test on the pooled error.
The standard error of the difference was the least significant difference divided by $t_{0.975,24}$.
For the 2024 to 2025 trials, data were available for each plot.
The recommendation was compared with standard practice by a Welch t-test at sites C1 and C2 and by a one-way ANOVA at sites H1, H2 and H3.
The protein response to N was estimated by linear regression on the seven ladder means.
The rate at which grain protein content reached 13\% was found by linear interpolation between adjacent ladder means.
Across site-seasons, mean differences between treatments were tested against zero by a one-sample t-test.
All t-tests were two-sided, and $P < 0.05$ was taken as significant.
Sample sizes are given in the figure legends.
Trigger counts from the in-season deployment are descriptive, and no statistical test was applied to them.

\subsection{Reporting summary}\label{sec:method-reporting}

Further information on research design is available in the Nature Portfolio Reporting Summary linked to this article.

\bmhead{Data availability}
The Broadbalk and Woburn Ley-Arable LTE data are openly available from the Electronic Rothamsted Archive under a CC-BY 4.0 licence \cite{Glendining2023Broadbalk,Glendining2022WoburnWheat,Glendining2022WoburnCropping}.
HadUK-Grid weather data are available from the CEDA Archive \cite{MetOffice2025HadUK}.
Derived curve tables, per-panel source data for all figures and tables, field-trial site means and the anonymised in-season trigger log are provided with this paper as Source Data.
Plot-level field-trial records and grower dialogue transcripts contain commercially sensitive information and are available from the corresponding author on reasonable request, subject to agreement with the participating farms.

\bmhead{Code availability}
The response-curve fitting, TabPFN recommendation model, rate regularisation and evaluation code are available from the corresponding author on request.
The bounding rules of the in-season layer are provided in the same way, the prompt templates are given as Supplementary Information, and all code will be deposited in a public repository on acceptance.

\bmhead{Acknowledgements}
The authors thank the participating growers and agronomists for their time and records during the deployment season.

\bmhead{Funding}
The authors disclose support for the research of this work from Innovate UK [grant number 10081615].

\bmhead{Author contributions}
X.W. and P.Y. conceived the study. X.W. developed the recommendation model, the rate regularisation and the in-season layer and drafted the manuscript. Y.Z., G.B. and M.Z. contributed to model development and analysis. S.J. and M.D. designed and ran the Harper Adams University field trials. S.M. designed and ran the commercial field trials and coordinated grower participation. P.H. advised on the long-term experiment data and agronomic interpretation. All authors reviewed and approved the manuscript.

\bmhead{Competing interests}
The authors declare no competing interests.

\bibliography{sn-bibliography}

\begin{thebibliography}{10}
\expandafter\ifx\csname url\endcsname\relax
  \def\url#1{\burl{#1}}\fi
\expandafter\ifx\csname urlprefix\endcsname\relax\def\urlprefix{URL }\fi
\providecommand{\bibinfo}[2]{#2}
\providecommand{\eprint}[2][]{\url{#2}}
\providecommand{\doi}[1]{\url{https://doi.org/#1}}
\bibcommenthead

\bibitem{vanGrinsven2022nitrogen}
\bibinfo{author}{van Grinsven, H. J.~M.} \emph{et~al.}
\newblock \bibinfo{title}{Establishing long-term nitrogen response of global
  cereals to assess sustainable fertilizer rates}.
\newblock \emph{\bibinfo{journal}{Nature Food}} \textbf{\bibinfo{volume}{3}},
  \bibinfo{pages}{122--132} (\bibinfo{year}{2022}).

\bibitem{Gong2024reactive}
\bibinfo{author}{Gong, C.} \emph{et~al.}
\newblock \bibinfo{title}{Global net climate effects of anthropogenic reactive
  nitrogen}.
\newblock \emph{\bibinfo{journal}{Nature}}  (\bibinfo{year}{2024}).

\bibitem{Galloway2008transformation}
\bibinfo{author}{Galloway, J.~N.} \emph{et~al.}
\newblock \bibinfo{title}{Transformation of the nitrogen cycle: Recent trends,
  questions, and potential solutions}.
\newblock \emph{\bibinfo{journal}{Science}} \textbf{\bibinfo{volume}{320}},
  \bibinfo{pages}{889--892} (\bibinfo{year}{2008}).

\bibitem{Menegat2022greenhouse}
\bibinfo{author}{Menegat, S.}, \bibinfo{author}{Ledo, A.} \&
  \bibinfo{author}{Tirado, R.}
\newblock \bibinfo{title}{Greenhouse gas emissions from global production and
  use of nitrogen synthetic fertilisers in agriculture}.
\newblock \emph{\bibinfo{journal}{Scientific Reports}}
  \textbf{\bibinfo{volume}{12}}, \bibinfo{pages}{14490} (\bibinfo{year}{2022}).

\bibitem{EU2020FarmFork}
\bibinfo{author}{{European Commission}}.
\newblock \bibinfo{title}{Farm to {Fork} strategy: For a fair, healthy and
  environmentally-friendly food system}.
\newblock \bibinfo{type}{Tech. Rep.}, \bibinfo{institution}{European
  Commission}, \bibinfo{address}{Brussels} (\bibinfo{year}{2020}).
\newblock
  \urlprefix\url{https://food.ec.europa.eu/horizontal-topics/farm-fork-strategy_en}.

\bibitem{UK2023EnvTargetsWater}
\bibinfo{author}{{Department for Environment, Food and Rural Affairs}}.
\newblock \bibinfo{title}{The {Environmental} {Targets} ({Water}) ({England})
  {Regulations} 2023}.
\newblock \bibinfo{howpublished}{Statutory Instrument 2023 No. 93}
  (\bibinfo{year}{2023}).
\newblock \urlprefix\url{https://www.legislation.gov.uk/uksi/2023/93/made}.
\newblock \bibinfo{note}{Accessed 3 November 2026}.

\bibitem{Xiao2024spatiotemporal}
\bibinfo{author}{Xiao, L.}, \bibinfo{author}{Wang, G.}, \bibinfo{author}{Wang,
  E.} \emph{et~al.}
\newblock \bibinfo{title}{Spatiotemporal co-optimization of agricultural
  management practices towards climate-smart crop production}.
\newblock \emph{\bibinfo{journal}{Nature Food}} \textbf{\bibinfo{volume}{5}},
  \bibinfo{pages}{59--71} (\bibinfo{year}{2024}).

\bibitem{defra2026cerealareas}
\bibinfo{author}{{Department for Environment, Food and Rural Affairs}}.
\newblock \bibinfo{title}{Cereal and oilseed areas in {England} at 1 {June}
  2026}.
\newblock \bibinfo{howpublished}{Accredited official statistics, GOV.UK}
  (\bibinfo{year}{2026}).
\newblock
  \urlprefix\url{https://www.gov.uk/government/statistics/cereal-and-oilseed-rape-areas-in-england/cereal-and-oilseed-areas-in-england-at-1-june-2025}.
\newblock \bibinfo{note}{Updated 27 August 2026. Accessed 5 November 2026}.

\bibitem{AHDB2021RB209}
\bibinfo{author}{{Agriculture and Horticulture Development Board}}.
\newblock \bibinfo{title}{Nutrient management guide ({RB209})}.
\newblock \bibinfo{type}{Tech. Rep.}, \bibinfo{institution}{Agriculture and
  Horticulture Development Board}, \bibinfo{address}{Kenilworth}
  (\bibinfo{year}{2021}).
\newblock
  \urlprefix\url{https://ahdb.org.uk/knowledge-library/rb209-section-1-principles-of-nutrient-management-and-fertiliser-use}.

\bibitem{Mi2025benchmark}
\bibinfo{author}{Mi, X.}, \bibinfo{author}{He, G.}, \bibinfo{author}{Qiu, W.}
  \emph{et~al.}
\newblock \bibinfo{title}{Data-driven nitrogen management benchmarks support
  {China}'s wheat self-sufficiency by 2030}.
\newblock \emph{\bibinfo{journal}{Nature Food}} \textbf{\bibinfo{volume}{6}},
  \bibinfo{pages}{692--702} (\bibinfo{year}{2025}).

\bibitem{Kindred2016autoN}
\bibinfo{author}{Kindred, D.~R.} \emph{et~al.}
\newblock \bibinfo{title}{Automating nitrogen fertiliser management for cereals
  ({Auto-N})}.
\newblock \bibinfo{type}{AHDB Project Report} \bibinfo{number}{561},
  \bibinfo{institution}{Agriculture and Horticulture Development Board}
  (\bibinfo{year}{2016}).
\newblock \urlprefix\url{https://repository.rothamsted.ac.uk/id/eprint/27215}.

\bibitem{Hawkesford2014}
\bibinfo{author}{Hawkesford, M.~J.}
\newblock \bibinfo{title}{Reducing the reliance on nitrogen fertilizer for
  wheat production}.
\newblock \emph{\bibinfo{journal}{Journal of Cereal Science}}
  \textbf{\bibinfo{volume}{59}}, \bibinfo{pages}{276--283}
  (\bibinfo{year}{2014}).

\bibitem{AHDB2026enduse}
\bibinfo{author}{{Agriculture and Horticulture Development Board}}.
\newblock \bibinfo{title}{End-use groups and specifications for milling wheat}.
\newblock \bibinfo{howpublished}{AHDB Knowledge Library}
  (\bibinfo{year}{2026}).
\newblock
  \urlprefix\url{https://ahdb.org.uk/knowledge-library/end-use-groups-and-specifications-for-milling-wheat}.
\newblock \bibinfo{note}{Accessed 5 November 2026}.

\bibitem{Keating2003APSIM}
\bibinfo{author}{Keating, B.~A.} \emph{et~al.}
\newblock \bibinfo{title}{An overview of {APSIM}, a model designed for farming
  systems simulation}.
\newblock \emph{\bibinfo{journal}{European Journal of Agronomy}}
  \textbf{\bibinfo{volume}{18}}, \bibinfo{pages}{267--288}
  (\bibinfo{year}{2003}).

\bibitem{Jones2003DSSAT}
\bibinfo{author}{Jones, J.~W.} \emph{et~al.}
\newblock \bibinfo{title}{The {DSSAT} cropping system model}.
\newblock \emph{\bibinfo{journal}{European Journal of Agronomy}}
  \textbf{\bibinfo{volume}{18}}, \bibinfo{pages}{235--265}
  (\bibinfo{year}{2003}).

\bibitem{Ruan2024datadriven}
\bibinfo{author}{Ruan, G.} \emph{et~al.}
\newblock \bibinfo{title}{Investigating data-driven approaches to optimize
  nitrogen recommendations for winter wheat}.
\newblock \emph{\bibinfo{journal}{Computers and Electronics in Agriculture}}
  \textbf{\bibinfo{volume}{220}}, \bibinfo{pages}{108857}
  (\bibinfo{year}{2024}).

\bibitem{Rose2016decision}
\bibinfo{author}{Rose, D.~C.} \emph{et~al.}
\newblock \bibinfo{title}{Decision support tools for agriculture: Towards
  effective design and delivery}.
\newblock \emph{\bibinfo{journal}{Agricultural Systems}}
  \textbf{\bibinfo{volume}{149}}, \bibinfo{pages}{165--174}
  (\bibinfo{year}{2016}).

\bibitem{GAO2024precision}
\bibinfo{author}{{U.S. Government Accountability Office}}.
\newblock \bibinfo{title}{Precision agriculture: Benefits and challenges of
  technology adoption for farms}.
\newblock \bibinfo{type}{Tech. Rep.} \bibinfo{number}{GAO-24-105962},
  \bibinfo{institution}{U.S. Government Accountability Office}
  (\bibinfo{year}{2024}).
\newblock \urlprefix\url{https://www.gao.gov/products/gao-24-105962}.
\newblock \bibinfo{note}{Accessed 2026}.

\bibitem{UK2015NVZ}
\bibinfo{author}{{UK Government}}.
\newblock \bibinfo{title}{The nitrate pollution prevention regulations 2015}.
\newblock \bibinfo{howpublished}{Statutory Instrument 2015 No. 668}
  (\bibinfo{year}{2015}).
\newblock
  \urlprefix\url{https://www.legislation.gov.uk/uksi/2015/668/contents}.
\newblock \bibinfo{note}{Accessed 2026}.

\bibitem{EA2026WaterFeb}
\bibinfo{author}{{Environment Agency}}.
\newblock \bibinfo{title}{Water situation: {February} 2026 summary}.
\newblock
  \bibinfo{howpublished}{\url{https://www.gov.uk/government/publications/water-situation-national-monthly-reports-for-england-2026/water-situation-february-2026-summary}}
  (\bibinfo{year}{2026}).
\newblock \bibinfo{note}{Accessed 5 April 2026}.

\bibitem{AHDB2026cropMarch}
\bibinfo{author}{{Agriculture and Horticulture Development Board}}.
\newblock \bibinfo{title}{Arable crop report says prospects are very good
  despite wet winter}.
\newblock \bibinfo{howpublished}{AHDB News} (\bibinfo{year}{2026}).
\newblock
  \urlprefix\url{https://ahdb.org.uk/news/arable-crop-report-says-prospects-are-very-good-despite-wet-winter}.
\newblock \bibinfo{note}{Published 31 March 2026. Accessed 6 November 2026}.

\bibitem{Kloppenburg1991social}
\bibinfo{author}{Kloppenburg, J., Jr.}
\newblock \bibinfo{title}{Social theory and the de/reconstruction of
  agricultural science: Local knowledge for an alternative agriculture}.
\newblock \emph{\bibinfo{journal}{Rural Sociology}}
  \textbf{\bibinfo{volume}{56}}, \bibinfo{pages}{519--548}
  (\bibinfo{year}{1991}).

\bibitem{Nonaka1994dynamic}
\bibinfo{author}{Nonaka, I.}
\newblock \bibinfo{title}{A dynamic theory of organizational knowledge
  creation}.
\newblock \emph{\bibinfo{journal}{Organization Science}}
  \textbf{\bibinfo{volume}{5}}, \bibinfo{pages}{14--37} (\bibinfo{year}{1994}).

\bibitem{Tzachor2023LLM}
\bibinfo{author}{Tzachor, A.} \emph{et~al.}
\newblock \bibinfo{title}{Large language models and agricultural extension
  services}.
\newblock \emph{\bibinfo{journal}{Nature Food}} \textbf{\bibinfo{volume}{4}},
  \bibinfo{pages}{941--948} (\bibinfo{year}{2023}).

\bibitem{Yi2025AgriLLM}
\bibinfo{author}{Yi, W.}, \bibinfo{author}{Zhang, L.}, \bibinfo{author}{Kuzmin,
  S.}, \bibinfo{author}{Gerasimov, I.} \& \bibinfo{author}{Liu, M.}
\newblock \bibinfo{title}{Agricultural large language model for standardized
  production of distinctive agricultural products}.
\newblock \emph{\bibinfo{journal}{Computers and Electronics in Agriculture}}
  \textbf{\bibinfo{volume}{234}}, \bibinfo{pages}{110218}
  (\bibinfo{year}{2025}).

\bibitem{Ji2023hallucination}
\bibinfo{author}{Ji, Z.} \emph{et~al.}
\newblock \bibinfo{title}{Survey of hallucination in natural language
  generation}.
\newblock \emph{\bibinfo{journal}{ACM Computing Surveys}}
  \textbf{\bibinfo{volume}{55}}, \bibinfo{pages}{1--38} (\bibinfo{year}{2023}).

\bibitem{Huang2025hallucination}
\bibinfo{author}{Huang, L.} \emph{et~al.}
\newblock \bibinfo{title}{A survey on hallucination in large language models:
  Principles, taxonomy, challenges, and open questions}.
\newblock \emph{\bibinfo{journal}{ACM Transactions on Information Systems}}
  \textbf{\bibinfo{volume}{43}}, \bibinfo{pages}{1--55} (\bibinfo{year}{2025}).

\bibitem{Frieder2023numerical}
\bibinfo{author}{Frieder, S.} \emph{et~al.}
\newblock \bibinfo{title}{Mathematical capabilities of {ChatGPT}}.
\newblock \emph{\bibinfo{journal}{Advances in Neural Information Processing
  Systems}} \textbf{\bibinfo{volume}{36}} (\bibinfo{year}{2023}).

\bibitem{Irisarri2026grassland}
\bibinfo{author}{Irisarri, J. G.~N.} \emph{et~al.}
\newblock \bibinfo{title}{Evaluating the transition of marginal grasslands to
  arable systems for human-protein production through an in silico long-term
  experiment}.
\newblock \emph{\bibinfo{journal}{Journal of Agriculture and Food Research}}
  \textbf{\bibinfo{volume}{28}}, \bibinfo{pages}{102980}
  (\bibinfo{year}{2026}).

\bibitem{deLara2023EONR}
\bibinfo{author}{de~Lara, A.}, \bibinfo{author}{Mieno, T.},
  \bibinfo{author}{Luck, J.~D.} \& \bibinfo{author}{Puntel, L.~A.}
\newblock \bibinfo{title}{Predicting site-specific economic optimal nitrogen
  rate using machine learning methods and on-farm precision experimentation}.
\newblock \emph{\bibinfo{journal}{Precision Agriculture}}
  \textbf{\bibinfo{volume}{24}}, \bibinfo{pages}{1792--1812}
  (\bibinfo{year}{2023}).

\bibitem{Wu2024agronomists}
\bibinfo{author}{Wu, J.} \emph{et~al.}
\newblock \bibinfo{title}{The new agronomists: Language models are experts in
  crop management}.
\newblock \bibinfo{howpublished}{arXiv preprint arXiv:2403.19839}
  (\bibinfo{year}{2024}).

\bibitem{Chen2025RLLLM}
\bibinfo{author}{Chen, D.} \& \bibinfo{author}{Huang, Y.}
\newblock \bibinfo{title}{Integrating reinforcement learning and large language
  models for crop production process management optimization and control
  through a new knowledge-based deep learning paradigm}.
\newblock \emph{\bibinfo{journal}{Computers and Electronics in Agriculture}}
  \textbf{\bibinfo{volume}{232}}, \bibinfo{pages}{110028}
  (\bibinfo{year}{2025}).

\bibitem{Balasubramaniam2026AgriSAGE}
\bibinfo{author}{Balasubramaniam, V.} \emph{et~al.}
\newblock \bibinfo{title}{{Agri-SAGE}: Simulation-grounded multi-agent {LLM}
  for context-aware agricultural advisory generation}.
\newblock \bibinfo{howpublished}{arXiv preprint arXiv:2607.00454}
  (\bibinfo{year}{2026}).

\bibitem{Chamara2026ChatGPT}
\bibinfo{author}{Chamara, N.} \emph{et~al.}
\newblock \bibinfo{title}{Can generative {AI} make farming decisions? current
  status and future pathways: a case study in row crop production with
  {ChatGPT}}.
\newblock \emph{\bibinfo{journal}{Artificial Intelligence in Agriculture}}
  \textbf{\bibinfo{volume}{16}}, \bibinfo{pages}{837--853}
  (\bibinfo{year}{2026}).

\bibitem{McCown2002changing}
\bibinfo{author}{McCown, R.~L.}
\newblock \bibinfo{title}{Changing systems for supporting farmers' decisions:
  Problems, paradigms, and prospects}.
\newblock \emph{\bibinfo{journal}{Agricultural Systems}}
  \textbf{\bibinfo{volume}{74}}, \bibinfo{pages}{179--220}
  (\bibinfo{year}{2002}).

\bibitem{Adereti2023barriers}
\bibinfo{author}{Adereti, D.~T.}, \bibinfo{author}{Gardezi, M.},
  \bibinfo{author}{Wang, T.} \& \bibinfo{author}{McMaine, J.}
\newblock \bibinfo{title}{Understanding farmers' engagement and barriers to
  machine learning-based intelligent agricultural decision support systems}.
\newblock \emph{\bibinfo{journal}{Agronomy Journal}}
  \textbf{\bibinfo{volume}{116}}, \bibinfo{pages}{1237--1249}
  (\bibinfo{year}{2023}).

\bibitem{Gardezi2026advisors}
\bibinfo{author}{Gardezi, M.} \emph{et~al.}
\newblock \bibinfo{title}{A socio-technical framework for analyzing crop
  advisors' preferences for {AI}-based decision support systems}.
\newblock \emph{\bibinfo{journal}{Technological Forecasting and Social Change}}
  \textbf{\bibinfo{volume}{226}}, \bibinfo{pages}{124601}
  (\bibinfo{year}{2026}).

\bibitem{Wang2026profit}
\bibinfo{author}{Wang, X.} \& \bibinfo{author}{Yang, P.}
\newblock \bibinfo{title}{Accurate prediction is not profitable advice:
  profit-based evaluation of machine learning nitrogen recommendations in
  winter wheat}.
\newblock \bibinfo{howpublished}{arXiv preprint arXiv:2608.27205}
  (\bibinfo{year}{2026}).

\bibitem{FAO2023FAOSTAT}
\bibinfo{author}{{Food and Agriculture Organization of the United Nations}}.
\newblock \bibinfo{title}{{FAOSTAT}: Crops and livestock products}.
\newblock \bibinfo{howpublished}{\url{https://www.fao.org/faostat/}}
  (\bibinfo{year}{2023}).
\newblock \bibinfo{note}{Accessed 2026}.

\bibitem{Glendining2023Broadbalk}
\bibinfo{author}{Glendining, M.} \& \bibinfo{author}{Poulton, P.}
\newblock \bibinfo{title}{Broadbalk wheat annual grain and straw yields
  1968--2022}.
\newblock \bibinfo{howpublished}{Electronic Rothamsted Archive, Rothamsted
  Research} (\bibinfo{year}{2023}).

\bibitem{Glendining2022WoburnWheat}
\bibinfo{author}{Glendining, M.}, \bibinfo{author}{Poulton, P.},
  \bibinfo{author}{Macdonald, A.}, \bibinfo{author}{MacLaren, C.} \&
  \bibinfo{author}{Clark, S.}
\newblock \bibinfo{title}{Woburn ley-arable experiment: yields of wheat as
  first test crop, 1976--2018}.
\newblock \bibinfo{howpublished}{Electronic Rothamsted Archive, Rothamsted
  Research} (\bibinfo{year}{2022}).

\bibitem{Glendining2022WoburnCropping}
\bibinfo{author}{Glendining, M.}, \bibinfo{author}{Poulton, P.},
  \bibinfo{author}{Macdonald, A.}, \bibinfo{author}{MacLaren, C.} \&
  \bibinfo{author}{Clark, S.}
\newblock \bibinfo{title}{Woburn ley-arable experiment cropping sequence
  1938--2020}.
\newblock \bibinfo{howpublished}{Electronic Rothamsted Archive, Rothamsted
  Research} (\bibinfo{year}{2022}).

\bibitem{MetOffice2025HadUK}
\bibinfo{author}{{Met Office}} \emph{et~al.}
\newblock \bibinfo{title}{{HadUK-Grid} gridded climate observations on a 1km
  grid over the {UK}, v1.3.1.ceda (1836--2024)}.
\newblock \bibinfo{howpublished}{NERC EDS Centre for Environmental Data
  Analysis} (\bibinfo{year}{2025}).
\newblock \bibinfo{note}{01 July 2025}.

\bibitem{Hollmann2025TabPFN}
\bibinfo{author}{Hollmann, N.} \emph{et~al.}
\newblock \bibinfo{title}{Accurate predictions on small data with a tabular
  foundation model}.
\newblock \emph{\bibinfo{journal}{Nature}} \textbf{\bibinfo{volume}{637}},
  \bibinfo{pages}{319--326} (\bibinfo{year}{2025}).

\bibitem{Hollmann2023TabPFN}
\bibinfo{author}{Hollmann, N.}, \bibinfo{author}{M{\"u}ller, S.},
  \bibinfo{author}{Eggensperger, K.} \& \bibinfo{author}{Hutter, F.}
\newblock \bibinfo{title}{{TabPFN}: A transformer that solves small tabular
  classification problems in a second}.
\newblock \emph{\bibinfo{journal}{International Conference on Learning
  Representations (ICLR)}}  (\bibinfo{year}{2023}).

\end{thebibliography}

\clearpage
\section*{Extended Data}
\setcounter{figure}{0}
\setcounter{table}{0}
\renewcommand{\figurename}{Extended Data Fig.}
\renewcommand{\tablename}{Extended Data Table}

\begin{table}[htbp]
\centering
\caption{The thirteen variants scored in Fig.~\ref{fig:pre-validation}g. Loss is the grain lost against the best possible N choice, in t/ha, as the mean over eight price ratios of the median over all 53 held-out Broadbalk LTE seasons. The baseline feeds the predicted anchors directly to the profit maximiser.}\label{tab:pre-variants}
\begin{tabularx}{\textwidth}{@{}l Y r r@{}}
\toprule
Group & Variant & Loss & Change \\
\midrule
Baseline & Predicted anchors to profit maximiser & 0.0803 & -- \\
Improve predictions & Conservative quantile anchors & 0.0890 & $+10.9\%$ \\
Improve predictions & Nitrogen-uptake feature stacking & 0.0757 & $-5.7\%$ \\
Improve predictions & Extra weather features & 0.0909 & $+13.2\%$ \\
Improve predictions & Larger ensemble, two seeds by eight & 0.0899 & $+12.0\%$ \\
Improve predictions & Median-anchor control, no model & 0.1071 & $+33.4\%$ \\
One-step regularisation & Shift only & 0.0821 & $+2.3\%$ \\
One-step regularisation & Shrink to global mean, single $\lambda$ & 0.0885 & $+10.3\%$ \\
One-step regularisation & Shrink to global mean, per-$\rho$ $\lambda$ & 0.0844 & $+5.1\%$ \\
One-step regularisation & Global shrink and shift & 0.0807 & $+0.6\%$ \\
One-step regularisation & Shrink to section mean only & 0.0924 & $+15.1\%$ \\
Full regularisation & Section shrink and shift & 0.0613 & $-23.7\%$ \\
Full regularisation & Section shrink, shift and uncertainty guard & \textbf{0.0609} & $\mathbf{-24.1\%}$ \\
\bottomrule
\end{tabularx}
\end{table}

\begin{table}[htbp]
\centering
\caption{Input features of the recommendation model. Weather windows are whole months and are aligned to the harvest year, so the winter window for harvest 1996 runs from October 1995 to February 1996. Weather values are taken from the HadUK-Grid 1 km dataset at the respective Rothamsted LTE grid points. Growing-season radiation is converted from monthly sunshine hours.}\label{tab:pre-features}
\begin{tabularx}{\textwidth}{@{}l l l Y@{}}
\toprule
Feature & Type & Unit & Window and source \\
\midrule
Previous crop & categorical & -- & Cropping records \\
Cultivar & categorical & -- & Experiment records \\
Soil type & categorical & -- & Site description \\
Section identifier & categorical & -- & Broadbalk section 0 to 9, off-site 999 \\
Winter rainfall & numeric & mm & October to February, HadUK-Grid \\
Growing-season rainfall & numeric & mm & March to July, HadUK-Grid \\
Growing-season radiation & numeric & MJ/m$^2$ & March to July, HadUK-Grid \\
Growing-season temperature & numeric & $^\circ$C, mean & March to July, HadUK-Grid \\
Grain-fill temperature & numeric & $^\circ$C, mean & June to July, HadUK-Grid \\
Grain-fill rainfall & numeric & mm & June to July, HadUK-Grid \\
\bottomrule
\end{tabularx}
\end{table}

\begin{figure}[htbp]
\centering
\includegraphics[width=\textwidth]{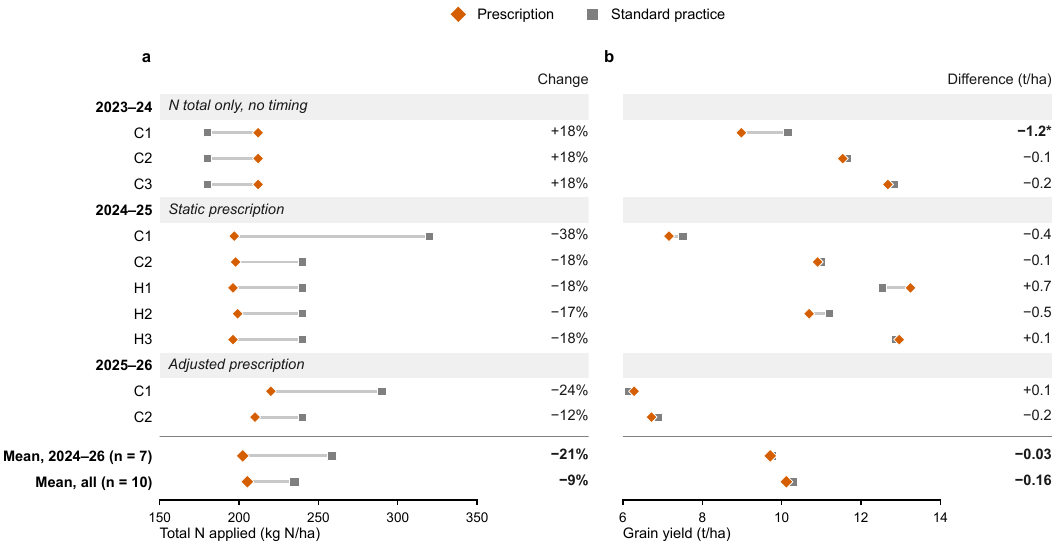}
\caption{\textbf{Nitrogen applied and grain yield under the recommendation and standard practice over three seasons.}
\textbf{a}, Total N applied.
\textbf{b}, Grain yield.
Rows are site-seasons, grouped by the form of the recommendation in each season.
In 2023 to 2024, only a seasonal N total was recommended, and doses were applied on the standard-practice dates.
Points show treatment means ($n = 4$ plots at C1 to C3, $n = 6$ at H1 to H3).
Asterisks mark significant differences ($P < 0.05$).
These were tested by Welch t-test or one-way ANOVA in 2024 to 2025, as in Fig.~\ref{fig:post-field-trials}, and by the trial LSD in other seasons.
The mean yield difference was not significant in 2024 to 2026 (one-sample t-test, $P = 0.86$) or across all seasons ($P = 0.31$).}
\label{fig:post-three-seasons-yield}
\end{figure}

\begin{figure}[htbp]
\centering
\includegraphics[width=\textwidth]{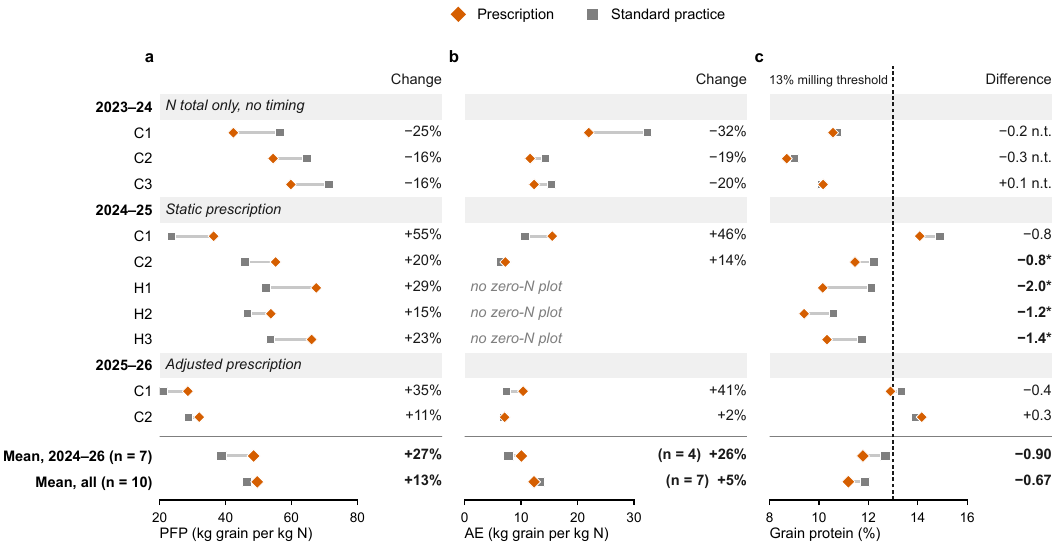}
\caption{\textbf{Yield per unit of nitrogen and grain protein over three seasons.}
\textbf{a}, Partial factor productivity (PFP), grain yield divided by total N applied.
\textbf{b}, Agronomic efficiency (AE), yield gain over the zero-N plot divided by total N applied.
AE is shown only where a zero-N plot was included.
\textbf{c}, Grain protein content.
The dashed line marks the 13\% milling-wheat threshold.
n.t., not tested.
Other conventions follow Extended Data Fig.~\ref{fig:post-three-seasons-yield}.}
\label{fig:post-three-seasons-nue}
\end{figure}

\begin{figure}[htbp]
\centering
\includegraphics[width=\textwidth]{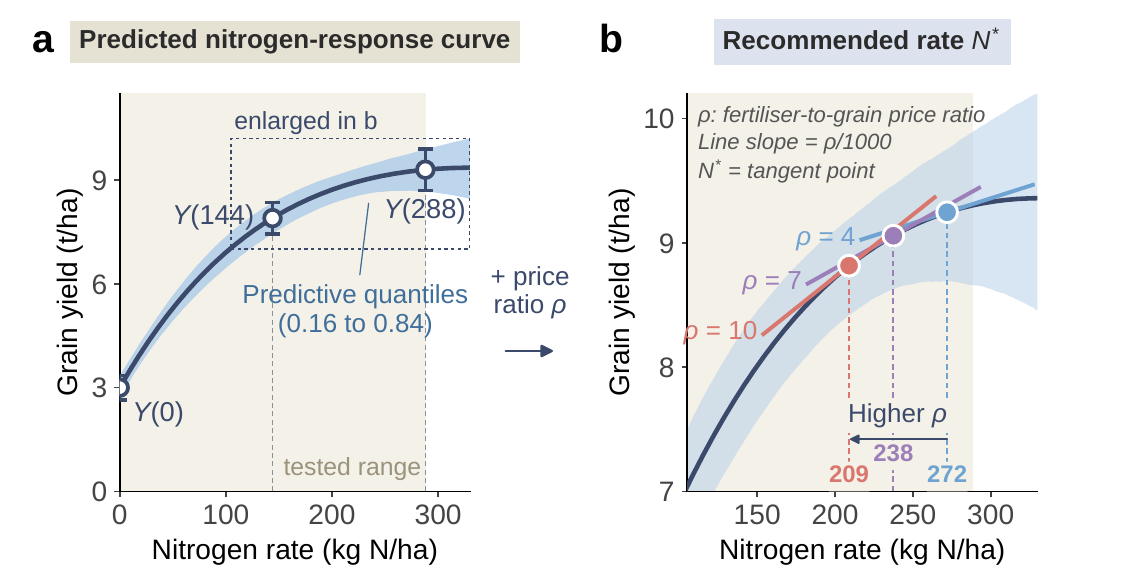}
\caption{\textbf{Illustration of the recommendation model.}
\textbf{a}, Nitrogen-response curve (equation~\eqref{eq:curve}) solved through the three predicted anchor yields, with the band between the 0.16 and 0.84 predictive quantiles.
\textbf{b}, Recommended rate $N^{*}$ (equation~\eqref{eq:argmax}) as the tangent point of a line of slope $\rho/1000$, for $\rho = 4$, 7 and 10.
Values are illustrative.}
\label{fig:method-nstar}
\end{figure}

\begin{figure}[htbp]
\centering
\includegraphics[width=0.8\textwidth]{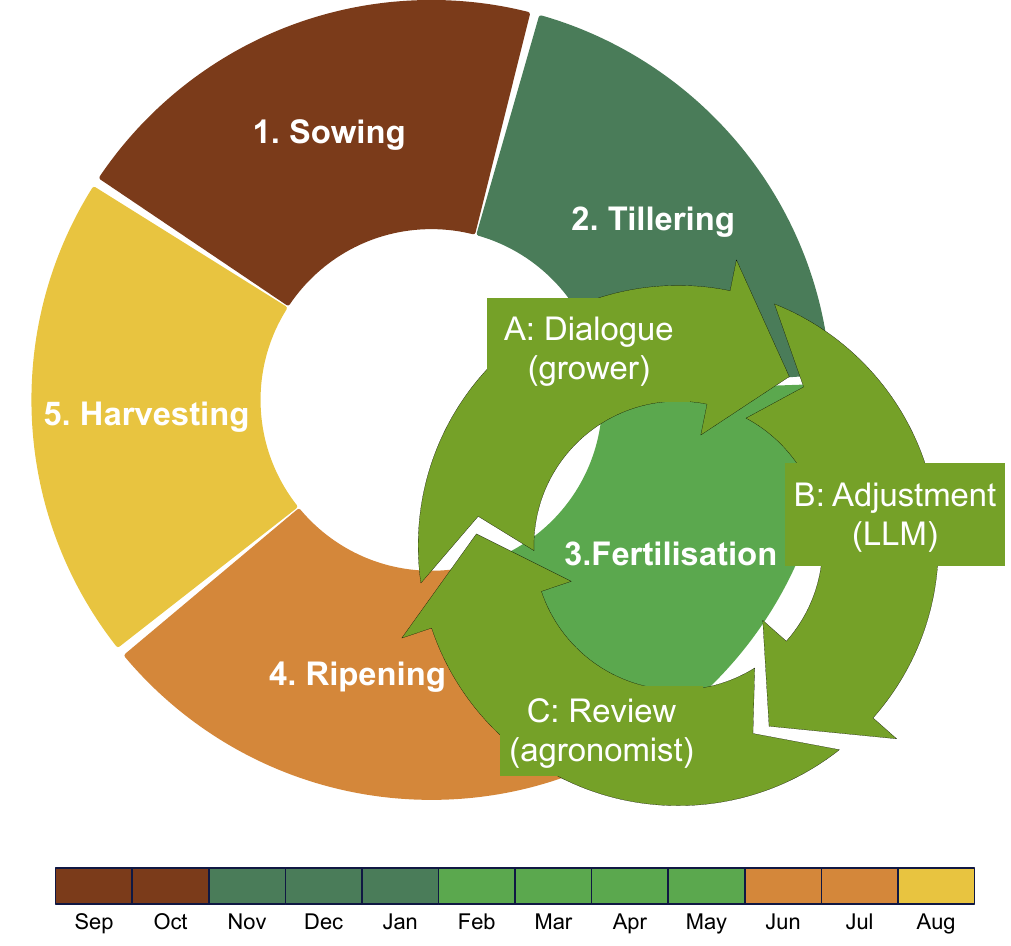}
\caption{Grower-in-the-loop workflow within the winter wheat season.
The outer ring shows the five crop stages, and the colour bar gives the months of each stage.
Within the fertilisation stage (February to May), the workflow links grower dialogue (A), LLM adjustment (B) and agronomist review (C).
The workflow is repeated adaptively as field conditions and grower constraints change.}
\label{fig:method-workflow}
\end{figure}

\end{document}